\pdfoutput=1

\documentclass[11pt]{article}

\usepackage[final]{acl}

\usepackage{times}
\usepackage{latexsym}

\usepackage[T1]{fontenc}

\usepackage[utf8]{inputenc}

\usepackage{microtype}

\usepackage{inconsolata}

\usepackage{placeins, amsmath, amssymb, amsthm, mathrsfs, calc,
  geometry, graphicx,
  txfonts, 
  ifthen, lscape,
  makeidx, manfnt,
  multicol, multirow,
  setspace, soul,
  tabto, booktabs, 
  textcomp,
  varioref, verbatim,
  wasysym, wrapfig,
  lipsum, eso-pic,
  eso-pic, fancyhdr,
  url, lineno,
  tikz, bm,
  float, subfig
}

\title{Local and Global Contexts for Conversation}

\author{Zuoquan Lin \and Xinyi Shen\\
Information and Computation Science Department\\
Peking University, Beijing, China\\
\texttt{\{linzuoquan,xinyi.shen\}@pku.edu.cn}
}

\begin{document}
\maketitle

\begin{abstract}
The context in conversation is the dialog history crucial for multi-turn dialogue. Learning from the relevant contexts in dialog history for grounded conversation is a challenging problem. Local context is the most neighbor and more sensitive to the subsequent response, and global context is relevant to a whole conversation far beyond neighboring utterances. Currently, pretrained transformer models for conversation challenge capturing the correlation and connection between local and global contexts. We introduce a \emph{local and global conversation model} (LGCM) for general-purpose conversation in 
open domain. It is a local-global hierarchical transformer model that excels at accurately discerning and assimilating the relevant contexts necessary for generating responses. It employs a local encoder to grasp the local context at the level of individual utterances and a global encoder to understand the broader context at the dialogue level. The seamless fusion of these locally and globally contextualized encodings ensures a comprehensive comprehension of the conversation. Experiments on popular datasets show that LGCM outperforms the existing conversation models on the performance of automatic metrics with significant margins.\footnote{Our codes are available at \url{https://github.com/PKUAI-LINGroup/LGCM}.} 
\end{abstract}

\section{Introduction}

\begin{figure*}[htb]
    \centering
    \includegraphics[width=1.0\textwidth]{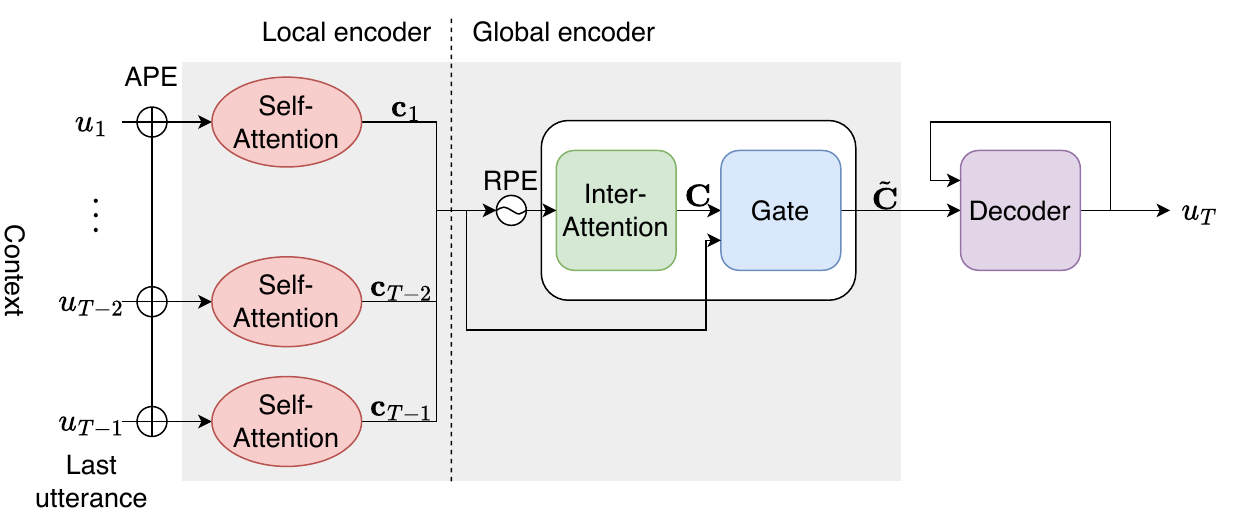}
    \caption{The architecture of LGCM: The encoder is hierarchical 
    attention consisting of the local and global encoders. The local encoders are standard transformer modules with PS (depicted the same color as Self-Attention) for each utterance in context. The global encoder consists of Inter-Attention and Gate for contextualized representations, which are sent to the cross-attention in the decoder. The decoder is a standard transformer decoder.
    }
    \label{fig:lgcm}
\end{figure*}

The role of context is significant in the similarity of words in a language. The contexts of a word are the neighboring tokens or grammatical structures. Contextualized embeddings encode both words and their contexts and generate contextualized representations. Language modeling captures distributed semantics embedded within these contextualized representations. The transformer-based pretrained language models (LMs) have become a foundation for NLP-like tasks \cite{Bommasani2021OnTO}. A well-established best practice in the field has consistently demonstrated that the utilization of large language models (LLMs) tends to yield superior performance in a wide range of NLP tasks, including conversational applications (say  \cite{wolf2019transfertransfo,adiwardana2020humanlike,roller-etal-2021-recipes,reed2021jurassic,thoppilan2022lamda}, among others).

\emph{Conversation models} (CMs) are generative sequence-sequence models for general-purpose conversations and learn the multi-agent distribution of utterances simultaneously. Most existing CMs are based on LMs, in which the LMs are used for accomplishing conversation by collaboration between agents that own their LMs or share a single LM in the spirit of parameter sharing (PS), where multiple models share the parameters in part or whole. In this paper, we consider the CMs with a single LM for two-agent conversation, such as human-machine dyadic dialogue.  

More specifically, CMs use either vanilla Transformer \cite{vaswani2017attention} as single-turn dialogue, such as question answering, where only the current utterance is considered as the history at any given turn, or for multi-turn dialogue adapt the Transformer architecture by concatenating multiple turns sequentially 
to capture the evolving context \cite{wolf2019transfertransfo,oluwatobi-mueller-2020-dlgnet,zhang-etal-2019-recosa}. Prominent examples of such CMs include 
TransferTransfo \cite{wolf2019transfertransfo}, Meena \cite{adiwardana2020humanlike}, Blender \cite{roller-etal-2021-recipes}, Athena \cite{reed2021jurassic} and LaMDA \cite{thoppilan2022lamda}, among others. 

The context in conversation is the dialog history crucial for multi-turn dialogue. CMs require an understanding of the dialog history, in the context of previous pairwise utterances and the current query at any turn. For example, as humans in everyday dialogue, the speaker's intent often cannot be detected by looking at the utterance level. In contrast, the speaker's acts are specific to each utterance and change throughout a whole dialog history at the dialogue level. One of the key challenges faced by CMs lies in striking the right balance between staying current which involves giving preference to recent utterances,  and drawing from the past effectively accumulating a prior understanding of the dialogue. The process of learning the relevant historical contexts necessary for fostering grounded and meaningful conversations remains a challenging problem in this domain. 

A criticism of the existing CMs is their inability to effectively utilize the available dialog history and gain a comprehensive view of a conversation \cite{sankar-etal-2019-neural}. A common problem of those CMs is their failure to establish meaningful correlations and connections between individual utterances. They often treat all the words as a single sequence and concatenate multiple turns in history into a single sequence, which neglects the distinct contexts of individual utterances within the broader dialogue history. 

To address the inherent problem of current CMs, we propose a more nuanced approach. In our model, we define each utterance as \emph{local context} for tokens at the utterance level and whole a dialogue as \emph{global context} for inter-utterances at the dialogue level. Moreover, we find it valuable to position the relationships among inter-utterances within a dialog history relative to one another. In our model, the conversation at different turns tells on each other, and all together, they tell what we talk about. 

Namely, we introduce a \emph{local and global CM} (LGCM) for multi-turn dialogue in open domain. It is a local-global hierarchical transformer model, illustrated in Figure~\ref{fig:lgcm}. It is an encoder-decoder architecture in which the decoder is the same as Transformer \cite{vaswani2017attention} with the cross-attention between the encoder and the decoder, but the encoder is a hierarchical attention structure. The encoder of LGCM consists of \emph{local encoders} and \emph{global encoder}. The local encoders are implemented by a standard transformer module (Self-Attention) for each utterance in the local context using absolute position encoding (APE). The global encoder consists of  \emph{Inter-Attention} and \emph{Gate} for contextualized representations in the global context, which are sent to the cross-attention in the decoder. The inter-attention is the attention between the current and all the utterances using relative positional encoding (RPE) \cite{shaw-etal-2018-self}. The gate fuses the representations of the local encoders and the inter-attention by a nonlinear transformation for local-global contextualized representation, see explanation in the subsection~\ref{sec:global}. 

In summary, the main contributions of this paper are the following: 
\begin{enumerate}
    \item[(1)] We are first trying to propose a CM that makes the connections between local context at the utterance level and global context at the dialogue level in a coherent way.
    \item[(2)] We propose a new attention mechanism (Inter-Attention) between current and historic utterances using RPE, which can separately deal with each utterance in a context. We extend the RPE from a single sequence in the self-attention to pairwise utterances within the conversation. 
\end{enumerate}

Experiments on popular datasets (DailyDialog, MultiWOZ, PersonaChat) show that LGCM takes advantage of the distinction between local and global contexts and outperforms the existing CMs on the performance of automatic metrics (PPL, BLEU, METEOR, NIST, ROUGEL) with significant margins (the best ratios range from 35.49\% to 71.61\%).

In the next section, we discuss the related works. In Section~\ref{sec:models}, we present LGCM in detail. In Section~\ref{sec:experiments}, we experiment on comparing LGCM with strong baseline CMs. Finally, we make some concluding remarks.

\section{Related works}\label{sec:works}

We concentrate on the CMs that use transformer-based LMs (see surveys \cite{tay2020efficient,de2021attention} for transformers and \cite{Bommasani2021OnTO} for LMs). Most CMs use LMs for multi-turn dialogue in open-domain \cite{wolf2019transfertransfo,adiwardana2020humanlike,roller-etal-2021-recipes,reed2021jurassic,thoppilan2022lamda}. SOTA CMs were large LMs (LLMs) trained specifically for conversation, such as ChatGPT\footnote{\url{https://chat.openai.com/}}, among other similar models.

Although LLMs can achieve the best practice from time to time, they scale up the Transformer, especially involving concatenating the dialog history into a single sequence. 
Small models are suitable for the study of CMs first, as the saying goes, it is difficult for a big ship to turn around.  Representative CMs are strong baselines based on small LMs such as GPT \cite{radford2018improving} and BERT \cite{devlin-etal-2019-bert}. Among them \cite{wolf2019transfertransfo,zhang-etal-2020-dialogpt,gu2020dialogbert,wu-etal-2020-tod,zhang2021dialoguebert}, TransferTransfo
\cite{wolf2019transfertransfo} trained especially on the basis of GPT, DialoGPT \cite{zhang-etal-2020-dialogpt} on GPT2 \cite{radford2019language}, and DialogBERT \cite{gu2020dialogbert} on BERT for dialog response generation.

Hierarchical encoders are a common framework for conversation. HRED was first introduced as two-level RNNs for multi-turn dialogue with a fuse between utterance and context dependencies \cite{sordoni2015hierarchical,serban2016building,serban2017multiresolution}. Most of the attention-based hierarchical models on multi-turn dialogue followed HRED architecture (say  \cite{xing2017hierarchical,tian-etal-2017-make,chen2018hierarchical,zhang-etal-2019-hibert,zhang-etal-2019-recosa,santra-etal-2021-hierarchical}, among others). Hierarchical CMs can have different mechanism designs \cite{zhu2018sdnet,Yang2019MakingHM,li2020dialbert}, some of which need an out-of-model mechanism such as learning-to-rank for ranking responses \cite{cao2007learning}, for instance, DialogBERT \cite{gu2020dialogbert}.  There was confusion about the performance between hierarchical versus non-hierarchical (i.e. single level) models. In \citet{lan2020kind}, hierarchical and non-hierarchical models for open-domain multi-turn dialog generation experienced: hierarchical models were worse than non-hierarchical ones, but hierarchical models with word-level attention were better than non-hierarchical ones. 
In \citet{santra-etal-2021-hierarchical}, it was claimed that hierarchical transformer models with context encoder are effective. Our work proves that hierarchical transformer models are better than non-hierarchical ones without any out-of-model mechanism. 

The effectiveness of combining local-global contexts was demonstrated in NLP and CV. It was effective to combine the benefits of using the attention for global context and using the CNN-like or the RNN-like for local context 
\cite{yang-etal-2016-hierarchical,zhang-etal-2019-recosa,gu2020dialogbert,wu2020lite,gulati2020conformer,wu2021cvt,peng2022branchformer}; or using the RNN-like for global context and using the attention for local context \cite{li2020dialbert}. In earlier works, hierarchical transformer encoders use only one token (say \verb+[CLS]+) as the hidden representation of sentence encoding to be fused in the context encoder (say HIBERT \cite{zhang-etal-2019-hibert}, DialogBERT \cite{gu2020dialogbert}). With the dominance of Transformer, it is natural to use Transformer to combine local-global contexts for sequence problems (say \cite{wu2021contextaware,santra-etal-2021-hierarchical,fang2022hierarchical,hatamizadeh2023global}, among others). HIER \cite{santra-etal-2021-hierarchical} is a strong baseline CM with hierarchical transformer encoders for individual utterances and context respectively, with some limitations compared to our model. In HIER, although contextual embeddings of all utterance tokens are input to the context encoder, the context is a concatenated sequence of utterances in a dialog history. In LGCM, we can separately deal with each utterance in a context and capture full contextualized representations of the local and global contexts by the attention and fuse mechanism. 

In essence, the concept of a hierarchical local-global architecture is not a novel one. However, what sets our model apart is our innovative approach to establishing meaningful correlations and connections between local and global contexts. We achieve this by introducing the Inter-attention and Gate mechanisms, which work in tandem to facilitate more coherent and contextually relevant conversations.

\section{Conversation models}\label{sec:models}

\subsection{Preliminaries}

We write $u = \{u_1, u_2, \cdots, u_T\}$ as a conversation with turn length $T \in \mathbb{N}$, where $\{u_{2k}\}_{k=1}^{\lfloor T/2 \rfloor}$ are utterances from one speaker and $\{u_{2k-1}\}_{k=1}^{\lceil T/2 \rceil}$ are those from the other speaker. We arrange that $u_T$ is the current response and $u_{T-1}$ is the last utterance. We introduce LGCM as an autoregressive generative model by the following equation of conditional distribution for the response $u_T$:

\begin{equation}
    P(u_T) = -\sum_{i=1}^{|u_T|} \log P(u_T^i | u_T^{<i}, u_{<T}; f_\theta),
    \label{eq:lgcm}
\end{equation}

\noindent where the conditional probabilities are computed by a neural network that is a (differentiable nonlinear) function $f_{\theta}$ with parameters $\theta$, which we shall take as a variant of Transformer \cite{vaswani2017attention}. The training objective is to maximize the average negative log-likelihood according to Equation~\ref{eq:lgcm}.

Recall that we distinguish local context for tokens in an utterance at the utterance level and global context for inter-utterances in a dialogue at the dialogue level. We encode local context for each utterance to capture more sensitive information from the neighboring tokens and global context for multiple utterances to capture inter-turn relevance from a dialog history. We obtain contextualized representations of utterances by fusing the local and global contexts. 

LGCM is implemented as a local-global encoder-decoder transformer (see Figure~\ref{fig:lgcm}). We modify the standard transformer encoder as local encoders with PS and global encoder and keep the decoder the same as the standard transformer decoder. 

\noindent \textbf{Embeddings}. Let $e(u_t^i)$ be a single token embedding (i.e. the $i$-th token in the $t$-th utterance), $e(u_t)$ an utterance embedding. We use APE for the token and utterance respectively. Let $p(i)$ be token positional embedding for the $i$-th token that is shared for each utterance and input in the local encoder, and $p_u(t)$ utterance positional embedding for the $t$-th utterance that is input in the global encoder. We use role embedding $r(t)$ for the $t$-th utterance to distinguish whether the speaker is a user or a bot. As usual, we use \verb+[bos]+and \verb+[eos]+ as the beginning and end of each utterance to separate between utterances.

We write $\bm{u}_t^i$ for input representation of token $u_t^i$ as follow:

\begin{equation}
    \bm{u}_t^i = e(u_t^i) + p(i) + r(t).
\end{equation}

\noindent What follows, we write $\bm{u}_t$ to denote the utterance embedding $e(u_t) = (\bm{u}_t^1, \cdots, \bm{u}_t^{|u_t|})$ for the sake of convenience. We share the input and output embedding matrices as usual done in past practice.

\noindent \textbf{Local encoder}. 
We use a standard transformer module as a local encoder of LGCM for each utterance in the local context. The transformer module is stacked layers of the multi-head self-attention followed by the feed-forward with layer normalization in a standard way. 
For each utterance $u_t$, an utterance representation $\bm{c}_t = \{\bm{c}_t^i\}_{i=1}^{|u_t|}$ is produced with the dimension of the value vector of $\bm{u}_t$, which is a context vector from a self-attention module. The locally contextualized representation $\bm{c}_t$ essentially summarizes the tokens in $\bm{u}_t$. 

For utterance embeddings $(\bm{u}_1, \cdots, \bm{u}_{T-1})$ in the context, the corresponding locally contextualized representations $(\bm{c}_1, \cdots, \bm{c}_{T-1})$ is the matrix of context vectors by grouping all the obtained context vectors together as columns. 

\noindent \textbf{Decoder}. We use a standard transformer decoder for LGCM. The decoder is stacked layers of the multi-head self-attention followed by the cross-attention with APE and the feed-forward with layer normalization in a standard way. 

\subsection{Global encoder}\label{sec:global}

We introduce a global encoder of LGCM at the dialogue level. The global encoder comprises the inter-attention and gate mechanism (Figure~\ref{fig:lgcm}). The hidden representations of the global encoder from the local contexts (Self-Attention) and the global context (Inter-Attention) are fused (via Gate) as the fully contextualized representations of the encoder of LGCM.

For locally contextualized matrix $\bm{c} = (\bm{c}_1, \cdots, \bm{c}_{T-1})$, we write globally contextualized representation as the matrix $\bm{C} = (\bm{C}_1, \cdots, \bm{C}_{T-1})$ correspondingly. The global representation $\bm{C}$ models the transformation of global context at the dialogue level from the local representation $\bm{c}$ at the utterance level as follows:

\begin{equation}
    \begin{aligned}
        \bm{C} = &\operatorname{LayerNorm}  (\operatorname{MultiHead}( \\
        &\operatorname{InterAttention} (\bm{c}, \bm{c}, \bm{c}) + \bm{c} )),
    \end{aligned}
\end{equation}

\noindent where $\operatorname{InterAttention} (Q, K, V)$ is the inter-attention mechanism as described in the following.  

\noindent \textbf{Inter-Attention}. We introduce the inter-attention to extend the attention mechanism to local-global inter-utterance attention by using RPE. The basic idea of InterAttention is that for any turn $t$, $\bm c_t$ attends to all the other $\bm c_s$s in the global context. Our RPE extends the original one \cite{shaw-etal-2018-self} from a single sequence in the self-attention to pairwise utterances for the conversation. We use RPE in attention not just for arbitrary pairwise token relations but also arbitrary pairwise utterance relations, which helps capture the structure of conversation in the sense that it refers to the relations between the tokens and utterances in input. 

$\operatorname{InterAttention}(Q,K,V)$ is defined according to the relation (relative distance) between the $t$-th utterance and the $s$-th utterance as input in the following: 

\begin{equation}
    \begin{aligned}
        \bm{A}_{t, s} & =\frac{1}{\sqrt{d_{o u t}}} \bm{c}_t \bm{W}^Q \left(\bm{c}_s \bm{W}^{K} + \bm{1}_{|u_s|} \bm{a}_{t, s}^K\right)^{\top}, \\
        \bm{C}_t & =\sum_{s=1}^{T-1}\operatorname{Softmax}\left(\bm{A}_{t, s}\right)\left(\bm{c}_s \bm{W}^{V}\right),
    \end{aligned}
\label{eq:inter-attention}
\end{equation}

\noindent where $\bm{W}^Q, \bm{W}^K, \bm{W}^V \in \mathbb{R}^{d_{in} \times d_{out}}$ are matrices to be learned for transforming $\bm{c}_t$, $\bm{c}_s$ to their $QKV$-representations, $\bm{a}_{t,s}^K \in \mathbb{R}^{d_{out}}$ is a learnable vector with the same dimension as 
$\bm{c}_s^j \bm{W}^K$ according to the relative distance between the $t$-th and the $s$-th utterances of the input. 
Namely, for a query $\bm{c}_t^i$, the inter-attention computes its globally contextualized representation over all the tokens, $\bm{c}_s^j$, belonging to their utterances that are locally contextualized representations in the following:

\begin{equation}
\begin{aligned}
\bm{C}_t^i &= \sum_{s=1}^{T-1} \sum_{j=1}^{|u_s|} \alpha_{t, s}^{i, j} (\bm{c}_s^j \bm{W}^{V}),\\
\alpha_{t, s}^{i, j} &= \operatorname{Softmax} (e_{t, s}^{i, j}),
\end{aligned}
\end{equation}

\noindent where $\alpha_{t, s}^{i, j}$ is the weight of $\bm{c}_t^i$ over $\bm{c}_s^j$. The logit $e_{t, s}^{i, j}$ is computed by the relative distance as follows:

\begin{equation}
    e_{t, s}^{i, j} = \frac{1}{\sqrt{d_{out}}} (\bm{c}_t^i {\bm{W}^{Q}}) (\bm{c}_s^j \bm{W}^{K} + a_{t,s}^K)^\top.
\end{equation}

\noindent Notice that we only take the relative distance representation for the key position, $\bm{a}_{t,s}^K$. As observed in past experiences \cite{shaw-etal-2018-self,huang-etal-2020-improve} and our ablation study, we observe that the key position encoding is key. 

In the original RPE, it is assumed that the relative position information is not useful beyond a certain distance and is clipped for the maximum relative position. We take the whole context length as the maximum; that is, we do not need to clip for it. Contrarily, we claim that the relative position information in a dialog history is useful for grounded conversation. The clipped maximum length possible does not allow the conversation to attend over an informative enough context. The global context depends on all the local contexts where information about the relative position representations selected by given attention heads is learnable. 

\noindent \textbf{Gate}. In the global encoder, the Gate follows from the inter-attention for the fusion of Self-Attention in the local context and Inter-Attention in the global context as fully contextualized representations. The fused encoding $\tilde{\bm{C}}$ is the fuse of the representation $\bm c$ of the local encoders and the one $\bm C$ of the inter-attention by a nonlinear transformation (Sigmoid) for local-global contextualized representation as follows:

\begin{equation}
    \begin{aligned}
        \bm{H} & =\operatorname{Sigmoid}([\bm{c} ; \bm{C}]\bm{W}), \\
        \tilde{\bm{C}} & =(1-\bm{H}) \odot \bm{C}+\bm{H} \odot \bm{c},
    \end{aligned}
    \label{eq:gate}
\end{equation}

\noindent where $\left[\bm{c}; \bm{C}\right]$ is the concatenation of $\bm{c}$ and $\bm{C}$, ${\bm W}$ is a learnable linear transformation, $\odot$ indicates element-wise (Hadamard) multiplication. Remember that the fused encoding $\tilde{\bm{C}}$ outputs to the cross-attention of the decoder.

Finally, a question may be asked whether the structure of LGCM for combining local-global contexts for more informative distribution brings up more computation burden than the Transformer. Most likely, we point out that the computational complexity of LGCM is less than Transformer. Let $L$ be the length of the input sequence and $d$ the dimension of the hidden state. The main computation burden for the single-head transformer encoder layer comes from matrix multiplications of self-attention and feed-forward network (FFN), namely $6Ld^2+4L^2d$ for self-attention and $16Ld^2$ for FFN, respectively. The local encoder of LGCM has the same structure as the Transformer encoder. The difference between them is that the local encoder of LGCM processes each utterance separately, while the Transformer encoder processes the concatenated sequence of utterances. Assume that the input sequence contains $N$ utterances with the same length the computation burden of the self-attention in the local encoder of LGCM is $6Ld^2+\frac{4L^2d}{N}$, which is more efficient than the Transformer encoder. For comparing the global encoder of LGCM and the Transformer encoder, we first consider the comparison between Inter-Attention and Self-Attention. As shown in Equation~\ref{eq:inter-attention}, the inter-attention adds a deviation about the relative distance to the key, which is negligible compared with matrix multiplication. Thus we consider that the computational complexity of the inter-attention and the self-attention is almost equal. We then consider the comparison between the Gate of LGCM and FFN. Since the computation burden of Sigmoid and element-wise multiplication can be ignored concerning matrix multiplication, the calculation amount of Gate is $4Ld^2$ according to Equation~\ref{eq:gate}, which is more efficient than FFN. To sum up, when the number of layers of both the LGCM encoder and the Transformer encoder is the same, the computational complexity of the LGCM encoder is less. This allows us to scale up the model to a large one.

\begin{table*}[htb]
\centering
\resizebox{\textwidth}{!}{
\begin{tabular}{@{}c|ccccccccccccccc@{}}
\toprule
\multirow{2}{*}{\textbf{Model}} & \multicolumn{5}{c}{DailyDialog}                                                                          & \multicolumn{5}{c}{MultiWOZ}                                                                             & \multicolumn{5}{c}{PersonaChat}                                                     \\ \cmidrule(l){2-16} 
                                & \textbf{PPL}   & \textbf{BLEU} & \textbf{METEOR} & \textbf{NIST}  & \multicolumn{1}{c|}{\textbf{ROUGEL}} & \textbf{PPL}  & \textbf{BLEU}  & \textbf{METEOR} & \textbf{NIST}  & \multicolumn{1}{c|}{\textbf{ROUGEL}} & \textbf{PPL}   & \textbf{BLEU} & \textbf{METEOR} & \textbf{NIST}  & \textbf{ROUGEL} \\ \midrule
Transformer                     & 30.03          & 6.86          & 10.61           & 26.48          & \multicolumn{1}{c|}{15.61}           & 5.01          & 12.95          & 22.05           & 63.62          & \multicolumn{1}{c|}{24.05}           & 36.66          & 7.65          & 10.52           & 40.95          & 15.77           \\
\midrule
TransferTransfo                 & 36.51          & 6.89          & 11.73           & 27.42          & \multicolumn{1}{c|}{17.11}           & 5.35          & 10.03          & 16.81           & 47.10          & \multicolumn{1}{c|}{19.48}           & 44.07          & 8.11          & 11.10           & 44.38          & 15.19           \\
DialoGPT                        & 42.90          & 7.36          & 12.78           & 29.04          & \multicolumn{1}{c|}{17.86}           & 5.25          & 12.59          & 21.24           & 61.75          & \multicolumn{1}{c|}{23.23}           & 40.74          & 7.74          & 10.38           & 41.58          & 15.21           \\
DialogBERT                      & 39.91          & 6.17          & 8.77            & 24.76          & \multicolumn{1}{c|}{11.35}           & 5.96          & 8.26           & 13.28           & 42.03          & \multicolumn{1}{c|}{14.51}           & 47.06          & 6.43          & 7.70            & 30.92          & 10.50           \\ \midrule
HIER                            & 27.89          & 6.70          & 11.47           & 25.12          & \multicolumn{1}{c|}{17.19}           & 5.05          & 13.06          & 22.15           & 64.62          & \multicolumn{1}{c|}{24.04}           & 37.42          & 7.75          & 10.31           & 41.81          & 15.52           \\ 
HIER-CLS                        & 30.34          & 6.57          & 11.19           & 25.26          & \multicolumn{1}{c|}{16.97}           & 5.05          & 12.92          & 21.62           & 65.86          & \multicolumn{1}{c|}{23.41}           & 39.38          & 7.91          & 10.68           & 43.60          & 15.69           \\ \midrule
LGCM                            & \textbf{26.48} & \textbf{8.36} & \textbf{14.08}  & \textbf{35.56} & \multicolumn{1}{c|}{\textbf{19.17}}  & \textbf{4.99}          & \textbf{13.26} & \textbf{22.79}  & \textbf{67.66} & \multicolumn{1}{c|}{\textbf{24.24}}  & \textbf{35.87}          & \textbf{8.41} & \textbf{11.79}  & \textbf{47.07} & \textbf{16.73}  \\ \bottomrule
\end{tabular}}
\caption{Automatic evaluation results on three datasets.}
\label{tab:automatic_results}
\end{table*}

\begin{table*}[htb]
\centering
\resizebox{\textwidth}{!}{
\begin{tabular}{c|ccccccccccccccc}
\hline
\multirow{2}{*}{\textbf{Model}} & \multicolumn{5}{c}{DailyDialog}                                                                          & \multicolumn{5}{c}{MultiWOZ}                                                                             & \multicolumn{5}{c}{PersonaChat}                                                     \\ \cline{2-16} 
                                & \textbf{PPL}   & \textbf{BLEU} & \textbf{METEOR} & \textbf{NIST}  & \multicolumn{1}{c|}{\textbf{ROUGEL}} & \textbf{PPL}  & \textbf{BLEU}  & \textbf{METEOR} & \textbf{NIST}  & \multicolumn{1}{c|}{\textbf{ROUGEL}} & \textbf{PPL}   & \textbf{BLEU} & \textbf{METEOR} & \textbf{NIST}  & \textbf{ROUGEL} \\ \hline
LGCM                            & \textbf{26.48} & \textbf{8.36} & \textbf{14.08}  & \textbf{35.56} & \multicolumn{1}{c|}{\textbf{19.17}}  & 4.99          & \textbf{13.26} & \textbf{22.79}  & \textbf{67.66} & \multicolumn{1}{c|}{\textbf{24.24}}  & 35.87          & \textbf{8.41} & \textbf{11.79}  & \textbf{47.07} & \textbf{16.73}  \\
-w/o IA                         & 26.87          & 7.74          & 13.45           & 32.14          & \multicolumn{1}{c|}{18.65}           & \textbf{4.98} & 13.15          & 22.24           & 65.83          & \multicolumn{1}{c|}{24.00}           & \textbf{35.63} & 7.85          & 10.52           & 43.03          & 15.08           \\
-w/o gate                       & 28.13          & 7.29          & 12.39           & 30.94          & \multicolumn{1}{c|}{17.29}           & 5.04          & 13.09          & 22.09           & 65.74          & \multicolumn{1}{c|}{24.00}           & 36.10          & 8.00          & 11.31           & 43.54          & 16.25           \\ \hline
\end{tabular}}
\caption{Ablation study results on Inter-Attention and Gate. `- w/o IA' refers to LGCM-w/o Inter-Attention, `- w/o Gate' refers to LGCM-w/o Gate.}
\label{tab:ablation_study}
\end{table*}

\section{Experiments}\label{sec:experiments}

\subsection{Setup}

\noindent\textbf{Datasets}. Experiments are conducted on three public-available English multi-turn dialog datasets as follows:

\begin{itemize}
    \item \emph{PersonaChat} \cite{zhang-etal-2018-personalizing}: This dataset is randomly paired and asked to get to know each other by chatting according to the given profiles, consisting of 164,356 utterances over 10,981 dialogs.
    \item \emph{DailyDialog} \cite{li-etal-2017-dailydialog}: This dataset covers a variety of topics in daily life, consisting of 
    102,979 utterances over 13,118 dialogs.
    \item \emph{MultiWoz} \cite{budzianowski-etal-2018-multiwoz}: This dataset comprises human-human written conversations in multiple domains and topics, consisting of 115,424 utterances over 8,438 dialogues. Although designed for task-oriented dialogue, the dataset is a good benchmark for open-domain response generation \cite{gu2020dialogbert}.   
\end{itemize}

\noindent \textbf{Comparison models}. We compare LGCM with baseline Transformer \cite{vaswani2017attention}, and four strong baseline CMs: TransferTransfo \cite{wolf2019transfertransfo}, DialoGPT \cite{zhang-etal-2020-dialogpt}, DialogBERT \cite{gu2020dialogbert} and HIER \cite{santra-etal-2021-hierarchical}. Both HIER and LGCM use hierarchical transformer encoders, the comparison between them demonstrates the effectiveness of the global encoder in our model. HIER-CLS \cite{santra-etal-2021-hierarchical} is a variant of HIER that takes a single token as the embedding for each utterance. We also include HIER-CLS for comparison.

When comparing models, we aim to eliminate the influence of pre-training data and model scale, focusing the comparison on model design. Hence, we re-implement these baseline models to match the scale of LGCM, and then train them on each dataset in a supervised manner.
Based on the characteristics of the baseline models, we divide them into two categories. The first group consists of Transformer, HIER, and HIER-CLS, which mainly differ from LGCM in the design of the encoder.
To directly reflect the effect of our designs in the LGCM encoder, for models in this group, we use the same input embedding and decoder as LGCM to eliminate the influence of irrelevant factors.\footnote{A subtle distinction is that since the Transformer lacks a hierarchical encoder structure, we add the utterance positional encoding in the input embedding when implementing the Transformer encoder.}
The models in the second group, DialoGPT, TransferTransfo, and DialogBERT, all have their special designs. For example, DialoGPT adopted a decoder-only structure, while TranferTransfo employs a multi-task learning paradigm. For these models, we make minimal modifications while retaining model-specific designs of the original models such as input embedding, multi-task learning, and decoding strategy.

\noindent\textbf{Implementation}. We use the transformers library to implement all the models \cite{wolf-etal-2020-transformers}.\footnote{\url{https://github.com/huggingface/transformers}} Transformer consists of 6 encoder layers and 6 decoder layers. All the hierarchical models (DialogBERT, HIER/HIER-CLS, and LGCM) consist of 3 local (or so-called utterance) encoder layers, 3 global (or so-called context) encoder layers, and 6 decoder layers. The decoder-only models (TransferTransfo and DialoGPT) consist of 6 decoder layers. The number of attention heads is 8, and the dimension of the hidden state is 512 for all the models. The maximum number of utterances allowed in the context is 7 \cite{adiwardana2020humanlike, gu2020dialogbert}.

The models are optimized by AdamW \cite{loshchilov2018decoupled}. The learning rate is tuned on the validation set, and the model checkpoints that performed best on the validation set are selected for testing. We adopt the sampling strategy for TransferTransfo and DialogBERT during generation as in the original papers. For the other models, we use greedy search.

\noindent\textbf{Metrics}. The models are evaluated by automatic evaluation metrics as follows:

\begin{itemize}
    \item \emph{Perplexity} is commonly used in NLP tasks, which measures the ability of a model to predict real samples. 
    \item \emph{BLEU} shows the $N$-gram similarity between the predicted results and the real ones \cite{papineni-etal-2002-bleu}. We present BLEU-4 in our experiments. 
    \item \emph{NIST} is an improved version of BLEU that takes into account the amount of information per $N$-gram \cite{doddington2002automatic}. 
    \item \emph{METOR} calculates recall in addition to precision and takes into account synonyms \cite{banerjee-lavie-2005-meteor}. 
    \item \emph{ROUGE-L} measures the similarity between the predicted text and the real one based on the longest common subsequence \cite{lin-2004-rouge}.
\end{itemize}

\subsection{Results}

\begin{figure*}[htb]
    \centering
    \subfloat[DailyDialog]{
    \includegraphics[width=0.32\textwidth]{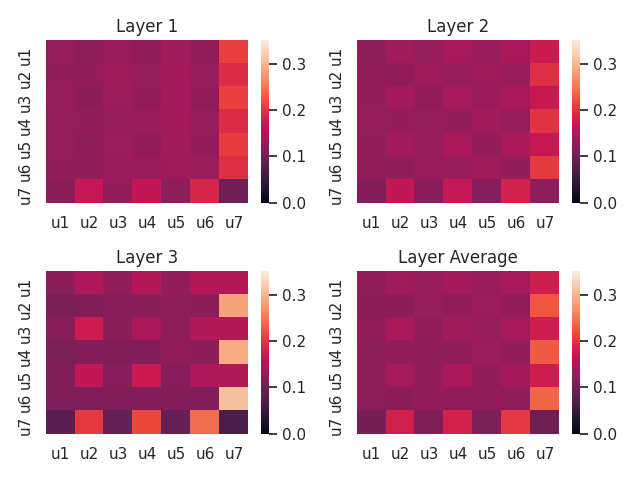}
    }
    \subfloat[MultiWOZ]{
    \includegraphics[width=0.32\textwidth]{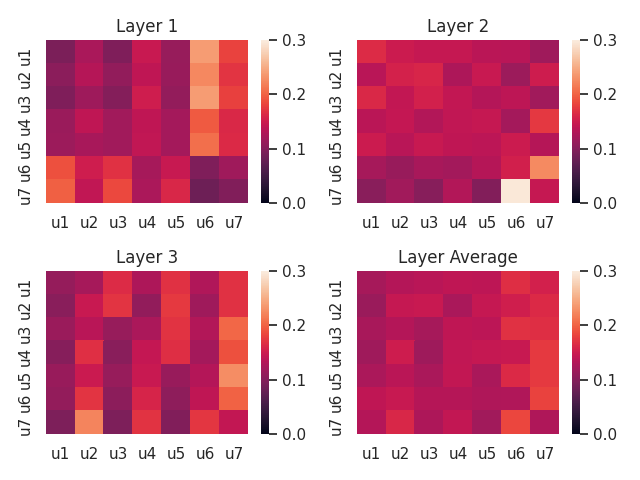}
    }
    \subfloat[PersonaChat]{
    \includegraphics[width=0.32\textwidth]{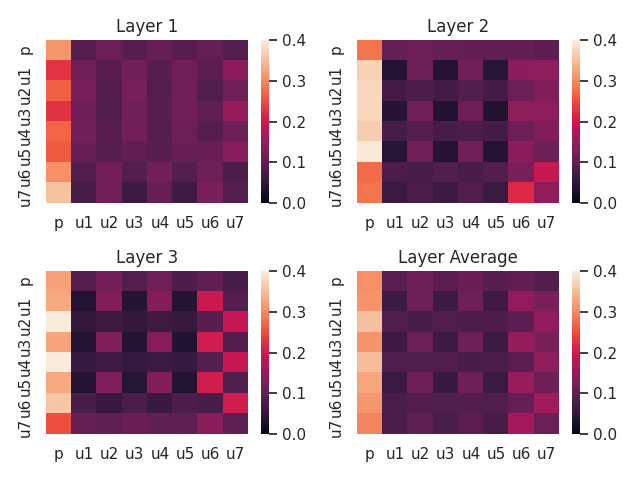}
    }
    \caption{The attention score visualization of the global encoder on the validation sets. The attention from $u_t$ to $u_s$ is calculated as $a_{t \rightarrow s} = \frac{1}{|u_t|} \sum_{i=1}^{|u_t|} \sum_{j=1}^{|u_s|} \alpha_{t,s}^{i,j}$.}
    \label{fig:glcm_attention_heatmap}
\end{figure*}

\begin{figure*}[htb]
    \centering
    \subfloat[DailyDialog]{
    \includegraphics[width=0.32\textwidth]{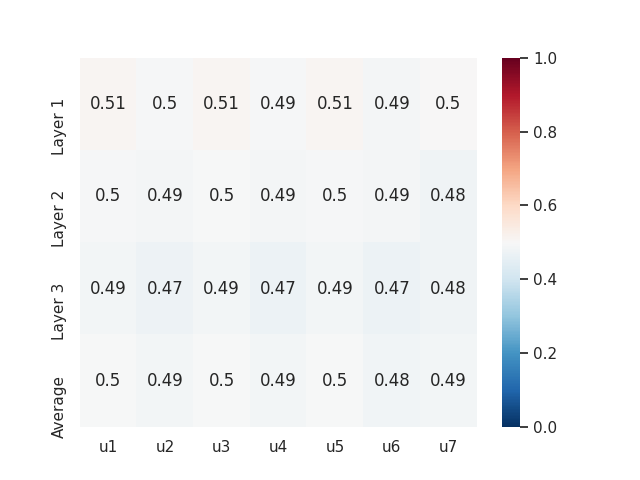}
    }
    \subfloat[MultiWOZ]{
    \includegraphics[width=0.32\textwidth]{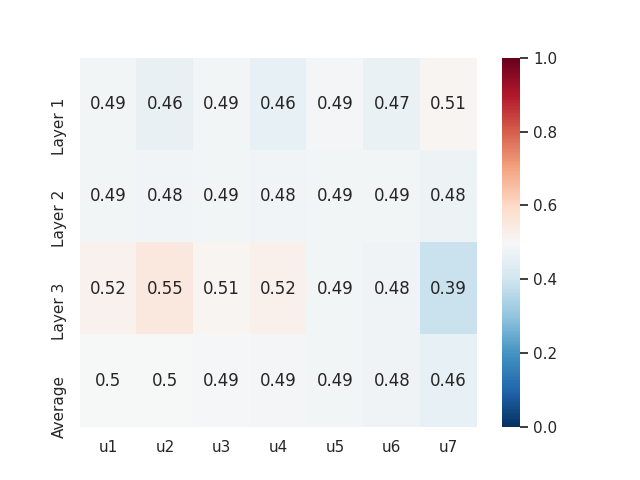}
    }
    \subfloat[PersonaChat]{
    \includegraphics[width=0.32\textwidth]{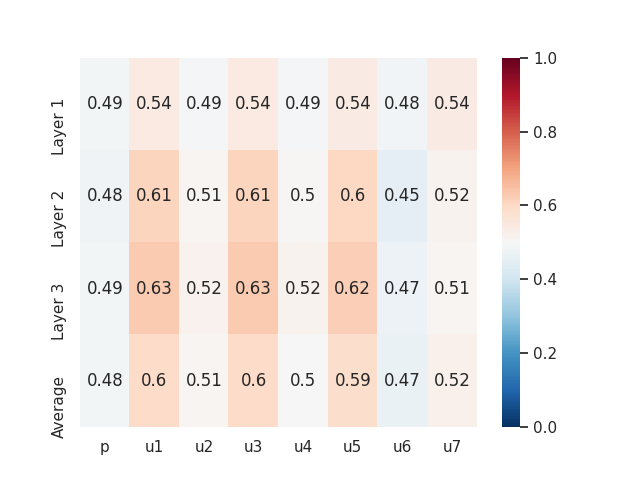}
    }
    \caption{The gate threshold visualization of the global encoder on validation sets. The values in the heatmap represent the proportion of the global information in the utterance representation, averaged across each token and each hidden dimension.}
    \label{fig:glcm_gate_heatmap}
\end{figure*}

\subsubsection{Evaluation}

The automatic evaluation results are shown in Table~\ref{tab:automatic_results}. We see that LGCM performs best on all the metrics with significant margins. The best ratios range from 35.49\% to 71.61\%, calculated from the table. The results show the effectiveness of LGCM through the fusion of local and global contexts. Therefore, we have positively answered that the distinction between local and global contexts is helpful in conversation.

\subsubsection{Ablation study}

To further examine the contributions of the two main designs in the global encoder of LGCM, we conduct ablation studies on Inter-Attention and Gate, respectively. To ensure the computing power of the model, when implementing LGCM-w/o Inter-Attention, we replace Inter-Attention with Self-attention, and when implementing LGCM-w/o Gate, we replace Gate with FFN.

As shown in Table~\ref{tab:ablation_study}, LGCM outperforms LGCM without Inter-Attention on DailyDialog. On the other two datasets, LGCM performs better than LGCM without Inter-Attention except for comparable to PPL. Additionally, removing Gate from LGCM results in a significant performance drop across all the metrics and all the datasets. This study shows that both Inter-Attention and Gate are the proper mechanisms for processing local and global contexts in conversation.

\subsection{Weight visualization}

To figure out how Inter-Attention and Gate help the model understand the contexts, we visualize the attention score and gate threshold in the global encoder of LGCM. 

Figure~\ref{fig:glcm_attention_heatmap} shows the heatmap of the attention weights between utterances. We see that the attention scores between utterances are greatly affected by the utterance's speaker.
For example, on the DailyDialog, the last utterance gives greater attention to utterances from partner utterances, especially at deeper layers. Furthermore, historic utterances tend to pay more attention to the latest utterances (the last two turns in our case), which is reasonable since the latest utterances are more relevant to the current dialog topic. In addition, all the historic utterances in PersonaChat have a high attention weight for the persona span, which reflects that the dialogs in the dataset are organized around the given profiles of both participants.

Figure~\ref{fig:glcm_gate_heatmap} shows the proportion of information from the global representations of utterances. We see that local and global contexts contribute considerably to the representations held among historic utterances and at different layers. This result demonstrates the necessity of using Gate to fuse local and global contexts dynamically. In addition, since Gate has reserved a considerable part of the information for each utterance, an utterance in the attention module usually pays more attention to the context other than itself, thus strengthening the inter-utterance interaction in the entire context.

\section{Conclusions}

Pretrained transformer models are adjusted by concatenating contexts into a single lengthy sequence. It is imperative to explore a variety of methods to encode the context effectively.

We have introduced a local and global conversation model for multi-turn dialogues in open domain. This model harnesses a hierarchical transformer encoder architecture, seamlessly integrating local and global contexts to enhance the efficacy of conversation. We have underscored the significance of distinguishing between the local context for tokens within an utterance at the utterance level and the global context for inter-utterances within a dialogue at the dialogue level. 
We hope that this study contributes to the comprehension of language models and conversational AI.

\section*{Limitations}

LGCM has some limitations. First, it is a small model with limited capability of conversation. We have not experienced scaling it up to a large one and pretraining it on big data. Second, we have not experienced extending it to the cases of multi-modal conversation and multi-task applications. These are areas where LGCM has not been applied, and they can be considered promising directions for future research.

\section*{Acknowledgements}
This work was supported by the National Natural Science Foundation of China under grant number 62076009.

\bibliography{anthology,custom}

\end{document}